# Morpheme Boundary Detection & Grammatical Feature Prediction for Gujarati : Dataset & Model

**Jatayu Baxi**
Dharmsinh Desai University / Nadiad
jatayubaxi.ce@ddu.ac.in

**Dr. Brijesh Bhatt**
Dharmsinh Desai University / Nadiad
brij.ce@ddu.ac.in

## Abstract

Developing Natural Language Processing resources for a low resource language is a challenging but essential task. In this paper, we present a Morphological Analyzer for Gujarati. We have used a Bi-Directional LSTM based approach to perform morpheme boundary detection and grammatical feature tagging. We have created a data set of Gujarati words with lemma and grammatical features. The Bi-LSTM based model of Morph Analyzer discussed in the paper handles the language morphology effectively without the knowledge of any hand-crafted suffix rules. To the best of our knowledge, this is the first dataset and morph analyzer model for the Gujarati language which performs both grammatical feature tagging and morpheme boundary detection tasks.

## 1 Introduction

As Natural Language Processing is increasingly becoming an active area of research with many important applications, most of the research is focused only on a few languages. Developing NLP tools for under resource languages is an essential task, as it not only opens considerable economic perspectives but also prevents its extinction and foster its expansion (Magueresse et al., 2020)

In this paper we present a morph analyzer for Gujarati. Gujarati is an Indo-Aryan language, spoken mainly in the Gujarat state of India. It is the 26th most widely spoken language with approximately 55 million speakers across the world. We identify various grammatical features of Gujarati morphology and prepare a gold data set of 16527 unique words. We have developed Bi-LSTM based morph analyzer inspired from (Premjith et al., 2018) and (Tkachenko and Sirts, 2018). For the training and evaluation of the morph analysis task, we have built the dataset for Gujarati language in the standard Unimorph format (Kirov et al., 2018). The neural architecture proposed in this paper does not require any language specific rules and captures linguistic characteristics of the language effectively.

The remaining of the paper is organized as follows : Section 2 describes related work. Section 3 describes the dataset details. Section 4 describes proposed approach. section 5 describes experiments and observations. Section 6 describes result analysis from the linguistic perspective. In section 7, we discuss conclusion and future research direction.

## 2 Related Work

Morphological analysis is the task of analyzing the structure of the morphemes in a word and is generally a prelude to further complex NLP tasks such as parsing, machine translation, semantic analysis. It is observed that for development of high end NLP tasks such as text summerization, question answering, NLI, machine translation systems, the word level utilities like morph analyzer, POS taggers can help in performance improvement.(Hedderich et al., 2021)(Pandya and Bhatt, 2021)(Parida and Motlicek, 2019)

Existing approaches to build a Morphological Analyzer can be broadly classified as Rule Based approaches and Machine Learning based approaches. Recent breakthroughs in the field of Deep Learning has motivated researchers to apply the neural models to the Morphological Analyzer problem.

Two-level morphology(Koskenniemi, 1984) was the first practical general model in the history of computational linguistics for the analysis of morphologically complex languages. The current C version two-level compiler, called TWOLC, was created at PARC(Beesley and Karttunen,

1992). (Kenneth R. Beesley and Lauri Karttunen, 2003) introduced XFST, a finite state morphology tool. Finite state transducer based approach have been used to develop Morph analyzer for many languages (Beesley, 1998)(Beesley, 2003)(Megerdoomian, 2004).(Kumar et al., 2012) developed a morphological analyzer for Hindi using this approach. Following Hindi, morphological generators were developed for other Indian languages e.g. Kannada (Melinamath and Mallikarjunmath, 2011), Oriya (Sahoo, 2003), etc. (Bharati et al., 2002) described another popular approach named Paradigm Based approach for morphological analysis. A Paradigm defines all the word forms that can be generated from given Root along with grammatical feature set.

Apart from the rule based techniques, there have been some efforts to develop machine learning based methods to develop morph analyzer for Indian languages. (Anand Kumar et al., 2010) have defined the morphological analysis problem as classification problem and experimented with various kernel methods to capture non linear relationships of the morphological features using SVM-Tool. (Srirampur et al., 2015) developed Statistical Morphological Analyzer for the Indian languages using linguistic features.

(Chakrabarty et al., 2016) proposed a Neural lemmatizer for the Bengali language. The proposed lemmatizer makes use of contextual information of the surface word to be lemmatized. (Heigold et al., 2016) investigated character based neural morphological tagger for morphologically rich languages having large tag set. They presented various neural architectures. The work is extended in (Heigold et al., 2017) by experimenting on 14 different languages. (Chakrabarty et al., 2017) introduced Deep Neural Network based method for lemmatization. This method works by identifying a correct edit tree for the word lemma transformation. (Premjith et al., 2018) Developed a Deep Learning approach for detecting morpheme boundaries. They studied agglutinative nature and sandhi splitting process of Malayalam language. (Tkachenko and Sirts, 2018) explore 3 different neural architectures named simple multi-class multilabel model, hierarchical model and sequence model for Morphological tagging. The encoder used for this work is same as the one used in (Lample et al., 2016). (Gupta et al., 2020) evaluated various neural morphological taggers for Sanskrit with the focus on the fact that good result for morphological tagging should be achieved without an extensive linguistic knowledge.

There have also been effort to develop unsupervised approaches for morphological tagging task. (J., 2005) described an Automatic Morphological Analyser which can be adopted for different languages. Morphessor (Creutz, 2005) is another widely used tool which performs unsupervised morphological segmentation. (Ak and Yildiz, 2012) and (Narasimhan et al., 2015) have also proposed unsupervised morphological analyzer using Trie based approach and Log linear methods.

To the best of our knowledge, very less work is reported in the area of developing a morph analyzer for the Gujarati language. (Patel et al., 2010) built a stemmer using handcrafted suffix list along with unsupervised learning. (Suba et al., 2011) built Hybrid Inflectional Stemmer and Rule-based Derivational Stemmer. (Baxi et al., 2015) developed rule based lemmatizer for Gujarati by hand crafting of suffix rules. The language independant models such as morfessor can not be used to develop full fledge morph analyzer as they only give morphological segmentation and do not perform morphological feature tagging. The model suggested by (Heigold et al., 2017) can not be directly used as the language specific training data for Gujarati language is not available in penn treebank dataset.

## 3 Gujarati morphology and Data set Generation

Gujarati is a verb-final language and has a relatively free word order, it is an inflectional language[1]. Words are formed by successfully adding suffixes to the root word in series. When suffixes are attached to the root, several morphophonemic changes take place. In this section, we describe the format and the details about the data set creation for the training and evaluation of morphological analyzer. For the creation of the dataset, we did a survey of available corpus for the Gujarati language. For the morphological analysis, it is preferable to have a POS tagged data, hence we have selected Gujarati Monolingual Text Corpus ILCI-II corpus for the creation of the dataset. The dataset is obtained from TDIL.[2]. We now discuss Gujarati morphology and details of dataset created for Noun, Verb and Adjective POS categories.

---

[1] https://en.wikipedia.org/wiki/Gujarati_grammar

### 3.1 Noun

Gujarati nouns participate in three genders and two numbers. The genders are masculine, feminine and neuter and the numbers are singular and plural. Gujarati nouns also inflect for various cases. Table 1 shows various cases with corresponding case markers. The data set for the noun category contains 6847 number of unique nouns. Along with each noun entry, the corresponding root form, gender, number and case information are marked manually.

| Case | Suffix |
|---|---|
| Nominative | φ |
| Genitive | નો,ની,નું,નાં ( *Nō,nī,nuṁ,nāṁ*) |
| Ergative | એ ( *ē*) |
| Objective/Dative | ને (*nē*) |
| Ablative | થી (*thī*) |
| Locative | માં (*māṁ*) |

Table 1: Case Markers for Gujarati Noun

### 3.2 Verb

Gujarati verbs inflect for gender, number, person, tense, aspect and mood features. Table 2 and Table 3 shows example of Gujarati verb with different moods and aspects respectively. The dataset for the verb category contains 6334 inflected verb forms. Each verb form is marked with corresponding root form and corresponding linguistic features.

### 3.3 Adjective

Gujarati adjectives can be classified in two types based on their nature of inflections. One class of adjectives do not inflect while the other class inflect for gender and number. Table 4 shows example of each category. The dataset for the adjective contains 3346 inflected adjective forms marked with linguistic features type, gender and number.

---
[2]Technology Development for Indian Languages (TDIL), http://http://tdil-dc.in

| POS Category | Features | Number of Words |
|---|---|---|
| Noun | Gender, Number, Case | 6847 |
| Verb | Gender, Number, Tense, Aspect, Person | 10128 |
| Adjective | Gender, Number | 3346 |

Table 5: Details about Dataset

## 4 Proposed Approach

We propose a morphological analyzer for Gujarati which performs morpheme boundary detection and grammatical feature tagging of a given inflected word. Gujarati is a morphologically rich language and manual hand crafting of rules is cumbersome process, hence we propose a deep learning based approach for this problem so that the features of an inflected word can be learned without supplying hand crafted rules. The morpheme boundary segmentation model inspired by (Premjith et al., 2018) and feature tagging module is based on the work reported in (Tkachenko and Sirts, 2018).

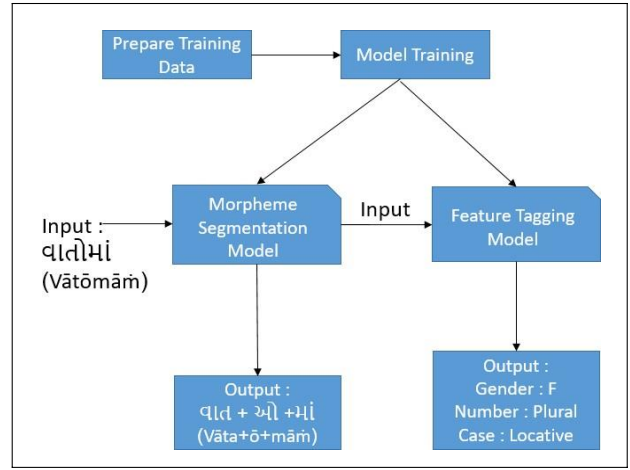

Figure 1: Block Diagram of the System

Figure 1 shows the overall architecture of the proposed system. We first prepare the training data and perform the model training for both tasks. The input word is passed to both models to get the corresponding morpheme segmentation and linguistic feature tagging outputs. For the preparation of training data, we represent inflected word as binary string and mark "1" in the position of the split character, rest all characters are marked as "0". Fig-ure 2 shows morpheme splitting example. Due to the rich morphological nature of Gujarati language, different word forms can be constructed by attaching various suffixes to the root word. Table 6 shows the sample output obtained for morpheme segmentation task.

| Mood | Example | Transliteration | English Translation |
|---|---|---|---|
| Indicative | ધન્વી ચોકલેટ ખાય છે. | Dhanvī cōkalēṭa khāya chē. | Dhanvi is eating a chocolate. |
| Imperative | સવારે વહેલો ઊઠજે. | Savārē vahēlō ūthajē. | Get up early in the morning.. |
| Conditional | જો હું ત્યાં હોત, તો હું તમને મદદ કરી શક્યો હોત. | Jō huṁ tyāṁ hōta, tō huṁ tamanē madada karī śakyō hōta. | Had I were there, I would have helped.. |
| subjunctive | એ અત્યારે દીપક ને ઘેર હોવાનો. | Ē atyārē dīpaka nē dhēra hōvānō. | He must be at Dipak's home right now. |

Table 2: Moods of Gujarati Verb

| Mood | Example | Transliteration | English equivilant |
|---|---|---|---|
| Simple | રામ અમદાવાદમાં રહે છે. | Rāma amadāvādamāṁ rahē chē. | Ram lives in Ahmedabad. |
| Progressive | રામ અત્યારે પુસ્તક વાંચી રહ્યો છે. | Rāma atyārē pustaka vāṅcī rahyō chē. | Ram is reading a book right now.. |
| Perfect | રામે પુસ્તક વાંચી લીધું. | Rāmē pustaka vāṅcī līdhuṁ. | Ram has finished reading a book. |
| Perfect Progressive | રામ સવારથી પૂજા કરી રહ્યો હતો. | Rāma savārathī pūjā karī rahyō hatō. | Ram was doing pooja since morning. |

Table 3: Gujarati Verb Aspects

| Type of Adjective | Example |
|---|---|
| Non-Inflected | ઉત્તમ (Uttama) |
| Inflected | સારો, સારી, સારું, સારા (sārō, sārī, sāruṁ, sārā) |

Table 4: Gujarati adjective inflection

Figure 2: Morpheme Splitting Example

| Inflected Word | Morpheme separation |
|---|---|
| વાતોમાં (Vātōmāṁ) | વાત (vāta) + ઓ (ō) + માં (māṁ) |
| પત્નીને (Patnīnē) | પત્ની (patnī) + ને (nē) |
| કરતો હતો (Karatō hatō) | કર (kara) + તો (tō) + હતો (hatō) |

Table 6: Morpheme Separation Example

In the Linguistic Feature Tagging module, we predict morphological features associated with an inflected word. Table 5 describes various features associated with different part of speech categories. We formulate this task as multi class classification problem. We have represented the class labels in a monolithic way such that each unique combination of different features is considered as one class. The number of classes for this task are 36, 198 and

13 for noun, verb and adjective categories respectively.

## 5 Experiments and Results

We use Keras Python library for the implementation of the system. Our model is sequential model with the first layer as embedding layer followed by a Bi-Directional LSTM layer followed by a dense layer for output prediction. We use Adam optimizer and binary cross entropy and categorical cross entropy loss for morpheme segmentation and feature tagging task respectively. We keep 80:20 ratio for training and testing of the model.Table 7 shows the results obtained for morpheme segmentation task.

| # of words in test set | Correctly segmented words | Accuracy |
|---|---|---|
| 4058 | 3614 | 89.05 |

Table 7: Morpheme boundary detection result - overall

Table 8 shows the results separately for each POS category.

| POS Category | # of Words | Correctly predicted words | Accuracy |
|---|---|---|---|
| Noun | 1369 | 1240 | 90.57 % |
| Verb | 2025 | 1761 | 86.96 % |
| Adjective | 669 | 645 | 97.49 % |

Table 8: Morphme Boundary Detection Results - POS category wise

To study the effect of POS category on the results, we repeat the experiments individually for each POS category. Table 9 shows the result of the morphological tagging task for various POS categories. We observe that system performs very well for morpheme boundary segmentation task across the POS categories.

We also compare the results of the neural morphological analyzer with an existing unsupervised morph analyzer Morphessor. Morphessor(Creutz, 2005) is a family of methods for unsupervised morphological segmentation. The first version of Morfessor, called Morfessor Baseline, was developed by Creutz and Lagus (2002) and its software implementation, Morfessor 1.0 was released by Creutz and Lagus (2005b). We have tested our dataset on morfessor implementation and compared the results of the neural model and the unsupervised model. For the Gujarati language, Morphessor implementation is available in Indic NLP(Kunchukuttan) library which is very popular NLP library for Indian Languages. However, the limitation of Morphessor is that it only performs morpheme segmentation task, whereas our proposed neural morphological analyzer performs both morpheme segmentation and morph feature tagging tasks. Due to this limitation, we are able to compare results for neural and unsupervised morph analyzer for only the morpheme segmentation task. We observe that the neural morphological analyzer outperforms unsupervised model by a large margin.

| | Accuracy in % | |
|---|---|---|
| POS Category | Neural Model | Unsupervised Model |
| Noun | 90.57 | 68.27 |
| Verb | 86.96 | 12.95 |
| Adjective | 97.49 | 25.72 |

Table 10: Accuracy comparison of neural and unsupervised model

## 6 Result analysis

### 6.1 Morpheme boundary detection

Using the LSTM based morpheme segmentation module, the system predicts a correct segmentation point for 3613 words out of 4058 total words in the test data set. Table 11 highlights few examples where system identifies correct splitting location for an inflected word:

Even though the morphemes splitting in all above cases are correct, It is observed that the first portion of the split may not be the valid root word every time. Table 12 highlights such examples.

We make an observation that the rules to form a valid root word are different for each word. These rules depend on POS category of the word and other grammatical features. Table 13 summarizes the rules.

| POS Category | Accuracy | Precision | Recall | F1-Score |
|---|---|---|---|---|
| Noun | 70.64 | 0.7 | 0.68 | 0.68 |
| Verb | 16.18 | 0.1 | 0.17 | 0.12 |
| Adjective | 85.85 | 0.78 | 0.61 | 0.68 |

Table 9: Morphological Feature Tagging Task Result

| Segmentation Example |
|---|
| પાત્રમાં (*Pātramāṁ*) → પાત્ર (*Pātra*)+માં (*Māṁ*) |
| દોડી (*Dōdī*) → દોડ (*dōḍa*) +ી (*ī*) |
| મગજને (*Magajanē*) → મગજ (*magaja*)+ને (*nē*) |
| દેખાશે (*Dēkhāśē*)→ દેખા (*dēkhā*)+ શે (*śē*) |
| યંત્રો (*Yantrō*) → યંત્ર (*yantra*)+ો (*ō*) |
| વ્યાજના (*Vyājanā*)→ વ્યાજ (*Vyāja*) + ના (*nā*) |
| છોકરા (*Chōkarā*) → છોકર (*chōkara*)+ા (*ā*) |
| ધંધાનું (*Dhandhānuṁ*)→ ધંધ(*dhandha*) + ાનું(*ānuṁ*) |
| ઈશારા(*Īśārā*)→ ઇશાર (*iśāra*) +ા(*ā*) |

Table 11: Segmentation Examples

| POS Category | Other Features | Rule |
|---|---|---|
| Noun / Adjective | Gender = Male | Attach Suffix -ો(*Ō*) |
| Noun / Adjective | Gender = Female | Attach Suffix -ી(*ī*) |
| Verb | - | Attach Suffix ું(*uṁ*) |

Table 13: Rules to form correct root word

We supply POS category of the word as an input to the system and obtain the grammatical features using morph tagging module. Using this information, accuracy of the morpheme boundary detection task can be further enhanced.

It is also observed that due to ambiguities in the word formation rules, in some cases, the system is not able to identify correct segmentation. For example, words વિદેશો (Vidēśō) is spitted correctly as વિદેશ +ો (Vidēśa + ō) and word જબરો (Jabarō) is spitted correctly as જબર +ો(Jabara + ō) . System tries to split the words ખુલાસો(Khulāsō) and કિનારો(Kinārō) using similar method leading to incorrect outputs ખુલાસ(Khulāsa) and કિનાર(Kināra). The issue here is that system considers ો (Ō) as the suffix but in some words ો(Ō) is part as the root word not as a suffix. The similar issue is observed in many other inflected words ending with suffix ી(Ī) as highlighted in the table 14

We also observe that the system does not produce correct segmentation in some cases where multiple suffixes are attached. For example, the correct segmentation of the word કારખાનાનું(*Kārakhānānuṁ*) is કારખાન(Kārakhāna)+ા(Ā)+નું(*Nuṁ* ) but the system does not identify any segmentation in the given word.

### 6.2 Grammatical feature prediction task

In this section, we do the result analysis of the grammatical feature prediction task from the linguistic perspective. We perform this analysis individually for each part of the speech category.
**Noun:**

For Noun, we consider gender, number and case as morphological features. The model is trained in such a way that based on the inflections that a word takes, it predicts corresponding grammatical features. For most of the cases we have good correlation between suffix and grammatical features, but in some cases the correlation does not hold. Due to these exceptions, sometimes there is an error in the feature prediction task. Consider two noun examples બજારો(Bajārō) and દાયરો(Dāyarō). Both words take similar suffix but in word બજારો(Bajārō), the suffix indicates plural marker but for the word ડાયરો(Dāyarō), the suffix is part of the word itself and the word is not plural. Similarly the word ઘટના(Ghatanā) is tagged with genitive case marker due to ના(Nā) attachment but actually the suffix is part of the word.

| Inflected word | Root morpheme detected by the system | Actual root word |
|---|---|---|
| દેખાશે (Dēkhāśē) → દેખા (dēkhā) + શે (śē) | દેખા (dēkhā) | દેખા (dēkhā) + વુ (vuṁ) → દેખાવુ (Dēkhāvuṁ) |
| છોકરા (Chōkarā) → છોકર (chōkara) + ા (ā) | છોકર (chōkara) | છોકર (chōkara) + ુ (U) + ં (ṁ) → છોકરું (Chōkaruṁ) |
| ધંધાનું (Dhandhānuṁ) → ધંધ (dhandha) + ાનું (ānuṁ) | ધંધ (dhandha) | ધંધ (dhandha) + ો (Ō) → ધંધો (dhandhō) |
| ઈશારા (Īśārā) → ઈશાર (iśāra) + ા (ā) | ઈશાર (iśāra) | ઈશાર (iśāra) + ો (Ō) → ઈશારો (Īśārō) |

Table 12: Examples of incorrect root identification

| Segmentation | Remark |
|---|---|
| છોકરી (Chōkarī) → છોકર (Chōkara) + ી (I) | Correct Segmentation |
| ગણતરી (ganatarī) → ગણતર (ganatara) + ી (I) | Incorrect Segmentation |
| શ્રેણી (śrēṇī) → શ્રેણ (śrēṇa) + ી (I) | Incorrect Segmentation |

Table 14: Segmentation Analysis

**Verb:**
For the verb category, we consider gender, number, person, tense and aspect features. Due to different combinations of features, we get total 198 classes for tag prediction task. The accuracy of the prediction task for verb is poor due to large number of classes. It is also observed that for different combinations of the features, same verb form exists which makes classification task more difficult. Table 15 highlights such examples:

A possible solution to address the above issue is to look at the input at the sentence level rather than word level. When the sentence level input is taken, verb features becomes clear and unambiguous. For example, with reference to the examples 1 and 2 from the above table, by looking at only રમતો હતો(Ramatō hatō), the person feature is not clear but when we look at the whole sentence : રામ રમતો હતો(Rāma ramatō hatō), the person feature becomes unambiguous ( person = 3rd). Similarly, by looking at only રમતી હતી(Ramatī hatī), the number feature is not clear but when we look at the whole sentence:છોકરીઓ રમતી હતી(Chōkarīō ramatī hatī), the Number feature becomes unambiguous (Number=PL) .

**Adjective**
We consider the type of an adjective, gender and number as features for morph feature tagging of an adjective. Consider the adjective અજ્ઞાની(Ajñānī). As per the language specification, this adjective does not inflect with gender and number but by looking at ી(Ī) suffix, the system predicts it as inflecting type of adjective with female gender.

To summarize, we observe that linguistic issues such as stem to root word generation, attachment of multiple suffixes and ambiguity in suffix rules affects the performance of the system.

## 7 Conclusion and Future Scope

In this paper we have proposed a Bi-LSTM based morphological analyzer for the Gujarati language. We have prepared the dataset and evaluated the proposed system. The system effectively performs morpheme boundary detection and morphological feature tagging tasks. With the proposed system, morphological analysis of unknown inflected word can be performed without the knowledge of linguistic rules. We have done result analysis from the linguistic perspective. We also conclude that the proposed model performs better than the existing unsupervised model.

In future, we aim to expand the dataset, and implement other neural architectures such as seq2seq model. We also aim to study sentence level dependency for morphological analysis.

| Sr No | Features | Verb Form |
|---|---|---|
| 1 | Gender= Male , Number=SG, Person=1st , Tense=Past, Aspect=Progressive | રમતો હતો(*Ramatō hatō*) |
| 2 | Gender= Male , Number=SG, Person=3rd , Tense=Past, Aspect=Progressive | રમતો હતો(*Ramatō hatō*) |
| 3 | Gender=Female, Number=SG, Person=3rd, Tense=Past, Aspect= Progressive | રમતી હતી(*ramatī hatī*) |
| 4 | Gender=Female, Number=PL, Person=3rd, Tense=Past, Aspect= Progressive | રમતી હતી(*ramatī hatī*) |

Table 15: Ambiguity in verb form generation

## Acknowledgement

The PARAM Shavak HPC computer facility is used for some of our experiments. We are grateful to the Gujarat Council of Science and Technology (GUJCOST) for providing this facility to the institution so that deep learning studies are being carried out effectively.